\documentclass[letterpaper,10pt,conference]{ieeeconf}
\IEEEoverridecommandlockouts
\usepackage{cite}
\usepackage{amsmath,amssymb,booktabs,graphicx,xcolor,multirow,subcaption,caption}
\usepackage{dsfont} 
\usepackage[percent]{overpic}
\graphicspath{{figs/}{figs/real_plot_recovery/}}

\newcommand{\method}{AquaOrbit}
\title{\bf \method: Sim-to-Real Reinforcement Learning for Underwater Target Orbiting under Intermittent Visual Feedback}
\author{Kanzhong Yao$^{1}$, Jinyi Leng$^{1,2}$, Hao Zhang$^{1,3}$, Zhe Sun$^{1,*}$, and Xuelong Li$^{1,*}$%
\thanks{$^{1}$Institute of Artificial Intelligence (TeleAI), China Telecom, China.}%
\thanks{$^{2}$Harbin Engineering University, China.}%
\thanks{$^{3}$Tongji University, China.}%
\thanks{$^{*}$Corresponding authors: Zhe Sun (\texttt{sunzhe@nwpu.edu.cn}); Xuelong Li (\texttt{xuelong\_li@chinatelecom.cn}).}%
}
\begin{document}
\maketitle
\thispagestyle{empty}
\pagestyle{empty}

\begin{abstract}
Intermittent visual loss disrupts target-relative feedback during
underwater orbiting, making it difficult to maintain coordinated
motion and reacquire a moving target. We present \method{},
a reinforcement-learning controller with a recovery module for
underwater target orbiting under interrupted visual feedback.
 During detection loss, the recovery module
uses latched line-of-sight, roll, and depth references to support
stabilization and target reacquisition. We train the controller
in Isaac Sim with dynamics, observation, and vision-loss
randomization. Evaluated without retraining in Gazebo/ROS~2
under a different physics engine and perception perturbations,
\method{} completes 20/20 orbiting trials in each of the static-
and moving-target conditions on an unseen variable-depth 3-D
trajectory. In the moving-target condition, it reduces mean
line-of-sight error by approximately 46\% relative to a PID-based
visual servoing controller with recovery while maintaining
comparable path-tracking accuracy; removing the recovery module
reduces completion to 9/20. Zero-shot physical deployment with
fully onboard perception and control demonstrates elliptical,
figure-eight, and variable-depth circular trajectories, including
the latter two trajectory types absent from training. The robot
maintains attitude stability during manual occlusions lasting
up to 8\,s and reacquires the target within 2.5\,s in the
reported attitude-induced field-of-view loss events.
\end{abstract}

\section{Introduction}
Autonomous target-relative motion is a core capability for underwater inspection, monitoring, and intervention. Among such behaviors, orbiting, also termed target circumnavigation, is a useful motion primitive: by continuously moving around a target at a desired stand-off distance, a vehicle can acquire multi-aspect inspection views, improve observability for active target localization, and maintain persistent observation of mobile marine assets and animals \cite{yang2020,wolek2021,masmitja2022}. Unlike conventional leader--follower control, which regulates a fixed relative offset, orbiting requires sustained nonzero tangential motion while maintaining the desired distance from the target. 

Classical visual control approaches typically rely on explicit geometric models. Position-based visual servoing (PBVS) reconstructs the target's relative three-dimensional pose and uses this estimate to compute control commands, whereas image-based visual servoing (IBVS) regulates image features through an analytical image Jacobian \cite{gao2015,heshmati2020,liu2023}. Both approaches face challenges underwater: pose estimation can degrade with turbidity \cite{cap,sun2026extreme}, increasing range, and oblique tag viewing angles, while image-based feedback becomes unavailable when the target leaves the camera's field of view or its features cannot be detected \cite{yao2023,yuan2026positive}. These challenges are particularly relevant to moving-target orbiting, where changing viewpoints and target motion can repeatedly interrupt visual feedback.

Recent approaches, such as the virtual elastic tether \cite{yao2026}, address temporary line-of-sight loss in leader--follower scenarios. However, extending such 2.5D task-specific mechanisms to three-dimensional orbiting is challenging, because the desired target-relative position changes continuously throughout the maneuver. Learned policies have also demonstrated the feasibility of underwater vehicle control, including performance comparable to PID control in the tasks studied \cite{ijrr2024,shao2026aiflow}. Nevertheless, robustness to intermittent target visibility remains insufficiently explored in these studies \cite{marinegym,cai2025}. The controller must
therefore combine coordinated target-relative tracking with
recovery behavior when visual feedback becomes unavailable.
\begin{figure}[t]
\centering
\includegraphics[width=\columnwidth]{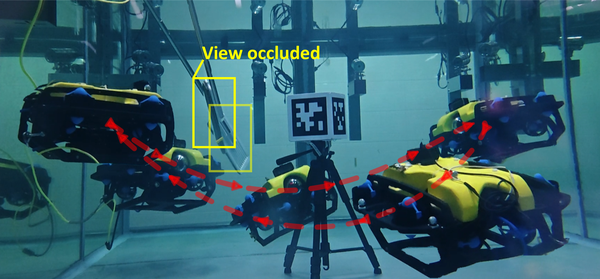}
\caption{3-D orbiting around the target AprilTag with continuous visual fixation on
its center. A manual occlusion of up to ${\sim}8$\,s is introduced from the
left.}
\label{fig:teaser}
\end{figure}

This motivates our central question: \emph{can a learned policy
use intermittent visual relative-position measurements and
onboard vehicle state to sustain orbiting around a moving
underwater target and recover from detection loss?}

We address this question with \method{}, a reinforcement-learning
(RL) controller for moving-target orbiting under intermittent
visual feedback~(Fig.~\ref{fig:teaser}). An AprilTag-based visual estimation node
\cite{olson2011} provides the target's three-dimensional position
relative to the camera. The policy combines this measurement
and its recent history with positional-reference features,
detection validity, onboard vehicle state, action feedback,
and recovery latch variables to generate individual thruster
commands. We formulate the task as a partially observable control problem to account for observation gaps caused by target motion, tag-face transitions, and detection failures. Training incorporates explicit vision-loss randomization, and a recovery module supports operation during intermittent target visibility. Policies are trained in Isaac Sim, evaluated without retraining in Gazebo to assess sensitivity to simulator-specific characteristics, and subsequently deployed on an underwater robot in an indoor test pond.

This paper makes the following contributions:
\begin{itemize}
\item An RL policy that drives a 6-DoF underwater vehicle to orbit a moving target from the target's relative 3D position and onboard proprioception alone, maintaining line-of-sight alignment while orbiting along unseen commanded trajectories.
\item Zero-shot deployment on a physical six-DoF vehicle with fully onboard perception and inference, sustaining orbits under intermittent visual dropout, external force disturbances, and irregular target motion, and merging into the orbit from a relaxed release region without a predefined guidance path.
\item A three-stage evaluation spanning Isaac Sim training, cross-simulator testing in Gazebo without retraining, and physical deployment in an indoor pond, using target-frame orbit metrics and comparisons against observation-matched and privileged baselines.
\end{itemize}

\section{Related Work}
\subsection{Reinforcement Learning for Underwater Vehicles}
Learning-based underwater control predates modern deep RL. Wettergreen
\emph{et al.} learned stable AUV controllers and coupled visual feature
tracking with vehicle guidance, demonstrating the early potential of
feature-driven learned control \cite{wettergreen1999}. Carlucho \emph{et
al.} mapped onboard sensory measurements to continuous low-level thruster
commands with an actor--critic policy and validated it on a real AUV
\cite{carlucho2018}. Deep RL has since been applied to goal-conditioned
underwater maneuvers: Anderlini \emph{et al.} compared DQN and DDPG with
classical controllers for simulated docking \cite{anderlini2019}, and Cai
\emph{et al.} trained six-DoF control policies in a parallelized
simulator and transferred them to a physical vehicle without real-world
fine-tuning \cite{cai2025}. Closest to our sensing setup, Bharti
\emph{et al.} combined TD3, PID demonstrations, and AprilTag localization
for sim-to-real AUV docking \cite{bharti2025}. Docking, however,
terminates at a goal configuration; orbiting instead demands persistent
tangential motion, continuous regulation relative to a \emph{moving}
target, and repeated recovery from field-of-view loss. 

\subsection{Underwater Visual Servoing}
Underwater visual servoing is commonly organized as PBVS, which controls
a reconstructed relative pose, or IBVS, which regulates features in the
image plane. Gao \emph{et al.} combined an estimated image Jacobian with
adaptive neural control for fixed-target dynamic positioning in
simulation \cite{gao2015}. Heshmati-Alamdari \emph{et al.} used visual
pose estimation with self-triggered nonlinear model predictive control,
validated experimentally on a small underwater vehicle
\cite{heshmati2020}. Moving targets add estimation and prediction
burdens: Liu \emph{et al.} proposed an IBVS tracker with an unscented
Kalman filter and image-Jacobian MPC, validated in simulation
\cite{liu2023}; Wang and Gao incorporated actor--critic adaptation into a
hybrid visual-servo controller for a vehicle--manipulator system, while
retaining a model-based kinematic layer and multicamera depth, in
simulation \cite{wang2024}. In contrast, AquaOrbit uses no analytical image Jacobian, model-based servoing law, or target-orientation estimate in its deployed control path: the learned policy maps the detector's relative-position output and onboard proprioception directly to thruster commands, and learns its own recovery behavior rather than switching to a hand-designed fallback.

\section{Problem Formulation}
\subsection{Moving-Target Orbiting}
Let $\mathcal{F}_T$, $\mathcal{F}_B$, and $\mathcal{F}_C$ denote the target,
robot's body, and camera frames. The task is to track a time-varying
\emph{guidance point} (a series of virtual references) $\mathbf{g}_t \in \mathbb{R}^3$ defined in the target
frame while keeping the target centered in the camera's field of view; the
guidance point decouples the desired viewpoint from the target body, so tracking it
drives the robot to orbit and inspect the target rather than approach it. The
reference orbit trajectory can take various predefined forms---vertical-plane
ellipses, figure-eights, and 3D orbits. Training uses only
parameter-randomized elliptical trajectories, while generalization to the
other trajectory types is assessed in the experimental section. At time
$t$, the policy must minimize the camera-frame tracking error
\begin{equation}
\mathbf{e}_{g,t} = \mathbf{e}_{\text{tag},t} + \mathbf{R}_{\mathcal{F}_C \leftarrow \mathcal{F}_T}\, \mathbf{g}_t,
\label{eq:guidance_error}
\end{equation}
where $\mathbf{e}_{\text{tag},t} = [e_{x,t}, e_{y,t}, e_{z,t}]^{\top}$ is the
observed target position in camera coordinates (from the AprilTag detector)
and $\mathbf{R}_{\mathcal{F}_C \leftarrow \mathcal{F}_T}$ rotates the
target-frame guidance-point offset into the camera frame. Notably, the target
frame $\mathcal{F}_T$ only translates with the target center and keeps its axes
parallel to the world frame without rotating with the target, so this rotation
is determined solely by the robot's attitude and is independent of the target's
orientation. Simultaneously, the policy must keep the target centered
in the image by driving the lateral and vertical components of
$\mathbf{e}_{\text{tag},t}$ toward zero:
\begin{equation}
\mathbf{e}_{\text{tag},t}^{\perp} = [e_{y,t},\, e_{z,t}]^{\top} \to \mathbf{0},
\label{eq:centering}
\end{equation}
where $e_{x,t}$ is the range along the optical axis and
$(e_{y,t}, e_{z,t})$ are the image-plane offsets.

\subsection{Observation Space}
\label{sec:obs}
The actor observes only quantities available in real-robot deployment: a visual target measurement $\mathbf{z}_t$ from the AprilTag estimation node, the IMU/AHRS vehicle state
$\mathbf{s}_t$, the applied thruster command $\mathbf{a}_t$, and
lost-target latch references $\mathbf{l}_t$. These form a single-frame
observation
$\mathbf{o}_t = [\mathbf{z}_t, \mathbf{s}_t, \mathbf{a}_t, \mathbf{l}_t]
\in \mathbb{R}^{44}$.

\textbf{Target measurement} $\mathbf{z}_t \in \mathbb{R}^{13}$: the
camera-frame target position $\mathbf{e}_{\text{tag}} \in \mathbb{R}^3$, the
guidance-point tracking error $\mathbf{e}_{g} \in \mathbb{R}^3$
(Eq.~\ref{eq:guidance_error}), a binary detection flag $\delta$ (set to $0$ when
the target leaves the FOV or is occluded), and the $10$- and $20$-step
displacements $\Delta_{10}\mathbf{e},\,\Delta_{20}\mathbf{e} \in \mathbb{R}^3$
of $\mathbf{e}_{\text{tag}}$, which capture short-horizon relative motion at
two timescales to trade off signal-to-noise ratio against timeliness.

\textbf{Vehicle state} $\mathbf{s}_t \in \mathbb{R}^{19}$ from the noisy
IMU/AHRS chain: attitude quaternion $\mathbf{q}$, angular velocity
$\boldsymbol{\omega}$, body-frame gravity direction, the body $x$/$z$ axes in
the world frame (a redundant, singularity-free attitude cue), and a $15$-frame
angular-velocity change $\Delta_{15}\boldsymbol{\omega}$. Linear velocity is
excluded, as underwater it can only be recovered by integrating IMU
acceleration, whose drift makes the estimate too unreliable.

\textbf{Applied action} $\mathbf{a}_t \in \mathbb{R}^{8}$ is the post-lag
per-thruster throttle. \textbf{Vision-loss latch} $\mathbf{l}_t \in
\mathbb{R}^{4}$ records the signed deviations of depth, heading, pitch, and
roll from references latched at the last detected frame ($\delta:\,1\!\to\!0$),
where the heading and pitch references are the line-of-sight attitude toward
the target's last-seen position (Sec.~\ref{sec:visionloss}).

\subsection{Action Space}
\label{sec:action}
The policy directly outputs $\mathbf{a}_t = \pi_\theta(\mathbf{o}_t) \in
[-1,1]^8$, the normalized throttle command for each of the eight thrusters.
The vectored eight-thruster configuration lets these commands span the full
six-DOF wrench, so all translational and rotational degrees of freedom are
learned jointly end-to-end. Each command passes through a thruster model
comprising a first-order throttle lag, a piecewise-linear throttle-to-speed
map with a deadband, and an empirical quadratic speed-to-thrust curve. 

\begin{figure*}[t]
\centering
\includegraphics[width=\textwidth]{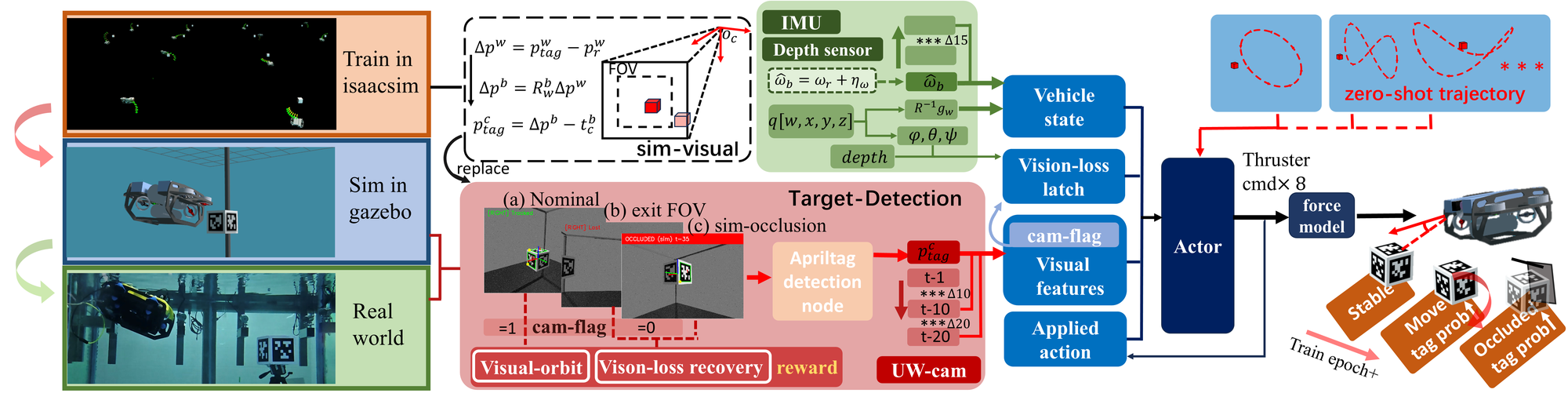}
\vspace{-6mm}
\caption{Framework of the AquaOrbit algorithm. Left: the sim-to-real pipeline from parallel training to real-world deployment. Middle: Gazebo simulation and real-world deployment rely on an upper-level target-position estimation module, whereas Isaac Sim training computes the camera-frame target position directly from ground-truth world-frame poses; the dashed box denotes the effective FOV. Right: curriculum learning progressively introduces static, dynamic, and occluded AprilTag targets to enhance robust target tracking.
}

\label{fig:framework}
\vspace{-4mm}
\end{figure*}

\section{Approach}
\subsection{Policy Architecture and Training}
\label{sec:policy}
The actor and critic are two separate three-layer MLPs (256 hidden units,
LeakyReLU activations, with layer normalization after each layer). We
train in 4096 parallel Isaac Sim environments (Fig.~\ref{fig:framework}); a single training run of
4000 iterations takes about one hour on a single NVIDIA RTX 4090. We adopt
an asymmetric actor--critic architecture: beyond the actor's observation,
the critic additionally receives privileged ground-truth simulator
information $\mathbf{o}_t^{\text{priv}} \in \mathbb{R}^{10}$, comprising the
camera-frame target position $\in \mathbb{R}^3$, the camera-frame target
tracking-point position $\in \mathbb{R}^3$, the body-frame linear velocity
$\in \mathbb{R}^3$, and the target FOV visibility $\in \mathbb{R}^1$; the
privileged visual information remains unaffected by target loss and random
occlusion.

Instead of rendering, we emulate the deployment-time AprilTag detector by
ground-truth geometry. Using the simulator's ground-truth vehicle and tag
poses, the tag center is mapped into the camera frame by a single rigid-body
transform,
\begin{equation}
\mathbf{p}_{\text{tag}}^{C}=\mathbf{R}_{W}^{B}\big(\mathbf{p}_{\text{tag}}^{W}-\mathbf{p}_{r}^{W}\big)-\mathbf{t}_{c}^{B},
\end{equation}
where $\mathbf{p}_{\text{tag}}^{W}$ and $\mathbf{p}_{r}^{W}$ are the tag and
vehicle positions in the world frame, $\mathbf{R}_{W}^{B}$ is the
world-to-body rotation from the vehicle attitude, and $\mathbf{t}_{c}^{B}$ is
the fixed camera offset in the body frame. We write
$\mathbf{e}_{\text{tag}}=\mathbf{p}_{\text{tag}}^{C}=[x,y,z]^{\top}$ ($x$ along
the optical axis, $y$ lateral, $z$ vertical). Since a real tag is
detected only when the \emph{whole} marker lies within the image, we bound it
by a sphere of radius $r$ and require the entire sphere to stay inside the
frustum, i.e.\ the center's distance to each lateral face is at least $r$:
\begin{equation}
\delta=\mathds{1}\!\left[\,x>0\;\wedge\;m_h\ge r\;\wedge\;m_v\ge r\;\wedge\;\|\mathbf{e}_{\text{tag}}\|<d_{\max}\,\right],
\end{equation}
with margins $m_h = x\sin(\theta_h/2)-|y|\cos(\theta_h/2)$ and
$m_v = x\sin(\theta_v/2)-|z|\cos(\theta_v/2)$. This yields an effective FOV that
is narrower than the nominal one and contracts as the tag approaches, and it
naturally induces a near blind zone; the slant-range term
$\|\mathbf{e}_{\text{tag}}\|<d_{\max}$ adds the far detection limit. We set
$r=L/\sqrt{2}$ (the circumradius of the $L=0.16$\,m tag face) and
$d_{\max}=5$\,m. Sensor noise, latency, and occlusion bursts
(Sec.~\ref{sec:randomization},~\ref{sec:visionloss}) are layered onto this
clean flag.

\subsection{Reward Design}
\label{sec:reward}
When the target is visible ($\delta_t{=}1$), the policy is driven by a
\emph{visual-orbit} reward: an instantaneous weighted sum of tracking, visual,
stability, and control groups,
\begin{equation}
\begin{aligned}
r_t = {} & w_{\text{tr}} r_t^{\text{track}}
+ w_{\text{ce}} r_t^{\text{center}}
+ w_{\text{at}} r_t^{\text{att}}
+ w_{\text{da}} r_t^{\text{damp}} \\
& + w_{\text{ac}} r_t^{\text{accel}}
+ w_{\text{ef}} r_t^{\text{eff}}
+ w_{\text{sm}} r_t^{\text{sm}},
\end{aligned}
\label{eq:reward_main}
\end{equation}
The tracking term applies a bi-exponential shaping on the camera-frame
guidance error, whose steep component gives tight regulation near the target
while the shallow one preserves the gradient at range.
The center term $r_t^{\text{center}}$ drives the image-plane error $(e_y,e_z)$ of
Eq.~\eqref{eq:centering} to zero, keeping the target centered in the FOV to
preserve observability; the attitude term applies a soft limit
$\sigma(x;k,c){=}(1{+}e^{k(x-c)})^{-1}$ that grants roll a permissive band,
letting the policy explore freer motion within a safe range, while pitch is
given a wider limit as it must keep the tag aligned across orbiting phases; and
the accel term
penalizes the angular acceleration $\Delta_{15}\omega_a$ over a $0.15$~s
window, mainly suppressing attitude oscillation/jitter around $0.5$--$2$~Hz
while leaving constant-rate functional maneuvers, whose acceleration is near
zero, unpenalized. The physical meaning, functional form, and coefficients of
all terms are summarized in Table~\ref{tab:reward}. During vision loss
($\delta_t{=}0$) the tracking and visual terms are replaced by a
\emph{vision-loss recovery} reward (Sec.~\ref{sec:visionloss}); the stability
and control terms remain active throughout.

\begin{table}[t]
\caption{Reward terms. The upper block applies when the target is visible
($\delta_t{=}1$); the lower block is the recovery reward during vision loss
($\delta_t{=}0$). $r{=}\|\mathbf{e}_{g}\|$, $\omega_a$ are body
angular rates, $u_i$ thruster throttles, $\sigma$ the soft-limit function, and
$(\cdot)_{\text{lk}}$ the value latched at the instant of loss.}
\label{tab:reward}
\centering
\footnotesize
\setlength{\tabcolsep}{3pt}
\begin{tabular}{llcl}\toprule
Group/term & Form & $w$ & Key params \\\midrule
\multicolumn{4}{l}{\emph{visual-orbit}} \\
track   & $\alpha_1 e^{-k_1 r}{+}\alpha_2 e^{-k_2 r}$ & $5.0$ & $\alpha{=}(.3,.7),k{=}(10,1)$ \\
center  & $\exp(-k_c\sqrt{e_y^2{+}e_z^2})$ & $3.0$ & $k_c{=}8$ \\
att     & $\tfrac12[\beta_r\sigma(|\phi|){+}\beta_p\sigma(|\theta|)]$ & $1.0$ & $k{=}(8,12),c{=}(.65,.85)$ \\
damp    & $\tfrac12\sum_a \gamma_a e^{-k_\omega\omega_a^2}$ & $1.0$ & $k_\omega{=}5,\gamma{=}(1.2,.5,.3)$ \\
accel   & $\tfrac12\sum_a \eta_a e^{-k_\alpha(\Delta_{15}\omega_a)^2}$ & $1.2$ & $k_\alpha{=}300,\eta{=}(.7,1,1)$ \\
eff     & $\tfrac18\sum_i e^{-k_e u_i^2}$ & $1.0$ & $k_e{=}3$ \\
sm      & $\exp(-k_s\|\mathbf{u}_t{-}\mathbf{u}_{t-1}\|)$ & $2.0$ & $k_s{=}1$ \\\midrule
\multicolumn{4}{l}{\emph{vision-loss recovery}} \\
pitch & $\exp(-k_p|\theta{-}\theta^{\star}_{\text{lk}}|)$ & $2.5$ & $k_p{=}6$ \\
roll  & $\exp(-k_r|\phi{-}\phi_{\text{lk}}|)$ & $2.5$ & $k_r{=}6$ \\
head  & $\exp(-k_h(1{-}\cos(\psi{-}\psi^{\star}_{\text{lk}})))$ & $2.0$ & $k_h{=}20$ \\
depth & $\exp(-k_d|z{-}z_{\text{lk}}|)$ & $3.0$ & $k_d{=}5$ \\
z-vel & $\exp(-k_v|\dot z|)$ & $2.6$ & $k_v{=}10$ \\
\bottomrule
\end{tabular}
\end{table}

An episode terminates on either range violation
($\|\mathbf{p}_{r}^{W}-\mathbf{p}_{\text{tag}}^{W}\|>3.0$~m, contact lost) or
sustained vision loss (target undetected for $>15$ consecutive frames,
counting only FOV departure and excluding transient occlusion).

  \begin{figure}[t]
    \centering
    \begin{overpic}[width=0.24\columnwidth]{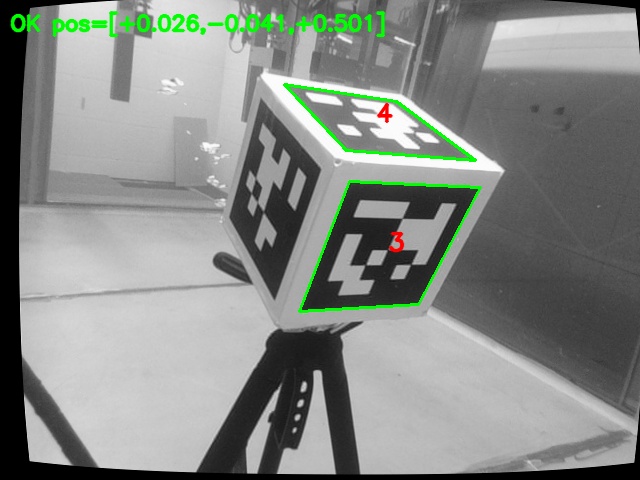}\put(80,60){\colorbox{white}{\small\bfseries 1}}\end{overpic}\hfill
    \begin{overpic}[width=0.24\columnwidth]{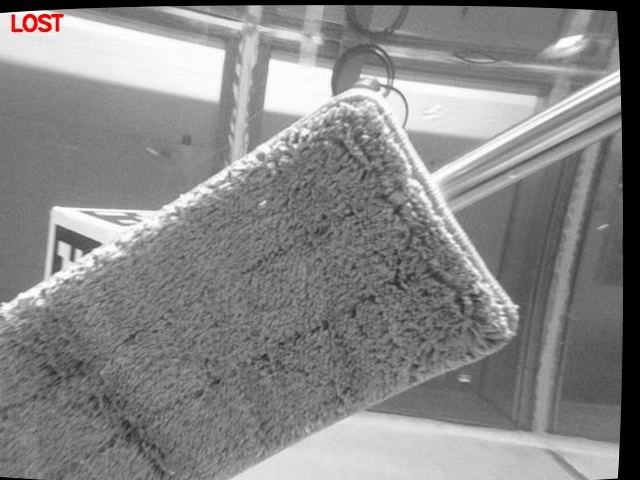}\put(80,60){\colorbox{white}{\small\bfseries 2}}\end{overpic}\hfill
    \begin{overpic}[width=0.24\columnwidth]{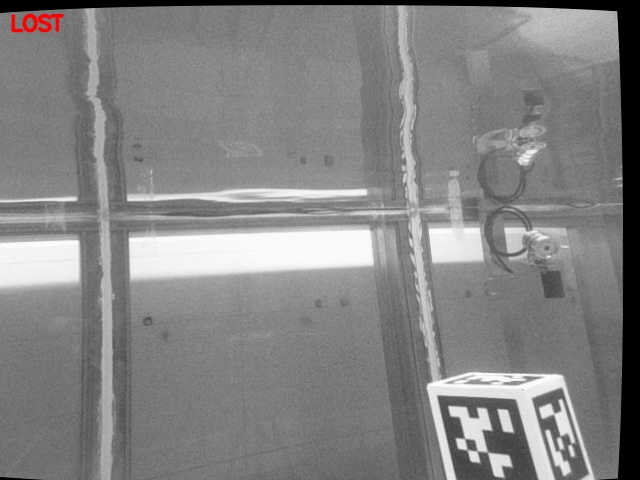}\put(80,60){\colorbox{white}{\small\bfseries 3}}\end{overpic}\hfill
    \begin{overpic}[width=0.24\columnwidth]{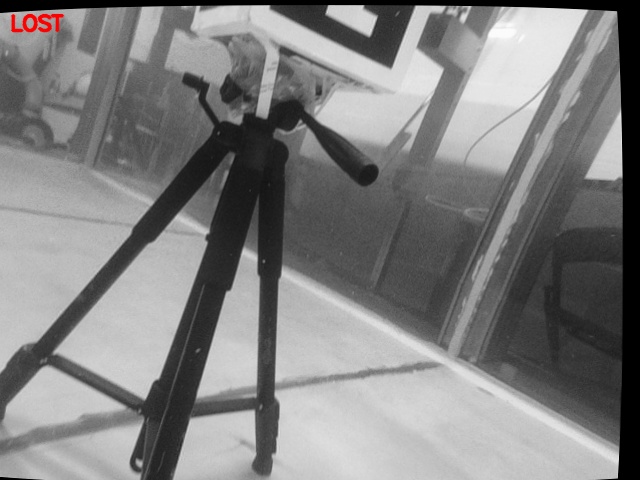}\put(80,60){\colorbox{white}{\small\bfseries 4}}\end{overpic}
    \caption{First-person views: (1) nominal detection, (2) occlusion, (3)--(4) partial FOV exit due to viewing angle.}
    \label{fig:visual-state}
    \vspace{-2mm}
  \end{figure}

\subsection{Vision-Loss Handling}
\label{sec:visionloss}
The policy must cope with two sources of detection loss encountered in
deployment (Fig.~\ref{fig:visual-state}): (i) \emph{geometric FOV
departure}, where the target leaves the field of view (FOV) because the
vehicle-to-target bearing grows too large---e.g.\ an abrupt change in
target velocity or a swing in the vehicle's own attitude; and (ii)
\emph{detection failure with the target still inside the FOV}, where the
target is transiently occluded by external moving obstacles or the
upstream vision module drops out briefly due to water turbidity or target
jitter. Both appear identically to the policy---as a single dropped
detection ($\delta_t{=}0$)---and both are handled by the same learned
recovery behavior.

When the target is lost ($\delta_t{=}0$), the replaced tracking, visual, and
attitude-limit terms form a \emph{vision-loss recovery} reward
\begin{equation}
r_t^{\text{hold}} =
w_p\, r_t^{\text{pitch}} + w_r\, r_t^{\text{roll}}
+ w_h\, r_t^{\text{head}}
+ w_d\, r_t^{\text{depth}} + w_z\, r_t^{\text{z\text{-}vel}},
\label{eq:reward_hold}
\end{equation}

\subsubsection{Latched Hold and Active Reacquisition}
At the instant of vision loss ($\delta:\,1\!\to\!0$), the depth, heading,
pitch, and roll references are latched and exposed to the policy through
$\mathbf{l}_t$ (Sec.~\ref{sec:obs}), and the recovery reward penalizes
deviation from them (lower block of Table~\ref{tab:reward}). Here heading and
pitch latch to the \emph{target-pointing} line-of-sight bearing on the last
visible frame, so the attitude terms both hold the camera on an occluded target
and actively reacquire a target lost to FOV departure. Roll latches its instantaneous body value to preserve
pose coherence. Horizontal position has no direct sensor, so no explicit term
is imposed---lateral motion is constrained indirectly by the control cost,
tolerating small-range coasting---while depth, being directly measurable, is
regulated by a depth-hold term.

\subsection{Domain and Observation Randomization}
\label{sec:randomization}
To bridge the sim-to-real gap, we follow the domain-randomization
paradigm~\cite{tobin2017,peng2018} and randomize sensor characteristics,
task geometry, and target motion at the start of every episode.
All ranges are listed in Table~\ref{tab:randomization}.

We inject noise matched to the onboard sensing chain: additive Gaussian
noise on the IMU angular rate, and an anisotropic, distance-dependent
term on the AprilTag detection. The visual observation is further passed
through a per-episode latency buffer (a fixed $3$--$5$ frame delay drawn
at reset) to emulate the processing and transmission lag of the tag
pipeline; only camera-detected quantities are delayed, while the
proprioceptive channels and the reward remain synchronous. Together these
force the policy to act robustly under noisy, delayed, and intermittent
feedback.

We further randomize the vehicle hydrodynamics, which are poorly identified
underwater. Starting from the nominal parameters of the open-source BlueROV2
benchmark simulator~\cite{bluerov2benchmark}, at each reset every environment
independently scales its added-mass, linear-damping, and quadratic-damping
matrices by a factor drawn from $\mathcal{U}[0.5, 1.5]$, and perturbs the
centre of buoyancy by a random 3-D offset. This exposes the policy to a wide spread of drag, inertial,
and restoring-torque behaviours, so it does not overfit the nominal model.

For task geometry, the reference orbit trajectory is an ellipse whose semi-axes,
plane tilt, angular rate, and initial phase are all randomized. We adopt a
three-stage curriculum that progressively increases task difficulty.

\textbf{(i) First 1000 iterations} The spawn pose is relaxed in three levels: the robot first
spawns along the ellipse's initial-phase direction, then at a
randomly sampled phase on the ellipse, and finally fully at random within
a cylindrical shell around the target ($360^\circ$ horizontal azimuth plus
a height offset of $[-0.35, 0.35]$~m). As the spawn distance and phase
offset relative to the trajectory's start point grow, the policy
progressively learns to approach distant trajectory points; under a large
phase offset the robot must aim at the tag and execute an arc-like motion
toward the target point to maximize reward. This capability substantially
relaxes the position and attitude requirements at the release point during
deployment and generalizes to other plausible scenarios.

\textbf{(ii) Iterations 1000--2000} The target
transitions from stationary to a Lissajous walk, with the walk probability
ramped from $0$ to $40\%$ and then held. As the target moves at a bounded but
unknown velocity, the policy leverages cross-frame jitter observations over 10-
and 20-frame windows to indirectly estimate its relative motion trend, yielding
smoother tracking of continuously moving targets under varying speed.

\textbf{(iii) After 2000 iterations} Since occlusion-type loss carries no
geometric signature, we overlay burst-mode detection dropout on the visual
observation to emulate the missed detections, tag occlusion, and tag-face
transitions of deployment. Its intensity (trigger probability and
maximum duration) is ramped linearly from $0$ over the following 500
iterations to its final values of $0.0025$ per step and $280$ frames,
forcing the policy to acquire a hold-and-reacquire behavior under
intermittent blindness and thereby become robust to the unstable visual
observations encountered underwater.

\begin{table}[t]
\caption{Domain randomization ranges (per episode unless noted; brackets
denote uniform $\mathcal{U}[\cdot]$; s.a.: semi-axis).}
\label{tab:randomization}
\centering
\footnotesize
\setlength{\tabcolsep}{3pt}
\begin{tabular}{llll}\toprule
Param. & Range & Param. & Range \\\midrule
IMU $\omega$ noise & $\sigma{=}0.008$~rad/s & Orbit s.a.\ (lat) & $[0.5,0.9]$~m \\
Tag noise (range) & $0.004{+}0.003d^2$ & Orbit s.a.\ (vert) & $[0.3,0.6]$~m \\
Tag noise (lat) & $0.002{+}0.001d$ & Orbit ang.\ rate & $[0.15,0.25]$~rad/s \\
Detect.\ dropout & 15--280 frames & Phase/azim. & $[0,2\pi]$ \\
Percept.\ latency & 3--5 frames & Plane pitch $\beta$ & $[{-}20,20]^\circ$ \\
Added-mass & $[0.5,1.5]{\times}$ & Init.\ dist. & $[0.5,1.4]$~m \\
Lin/quad damp. & $[0.5,1.5]{\times}$ & Spawn height & $[{-}0.35,0.35]$~m \\
CoB offset & $\pm0.3/0.3/0.5$~cm & Walk ampl. & $(2,2,1)$~m \\
Walk freq. & $[0.005,0.015]~hz$ & Walk phase & $[0,2\pi]$ \\
\bottomrule
\end{tabular}
\end{table}

\section{Experiments \& Results}
\subsection{Sim-to-Sim Cross-Engine Evaluation}
\label{sec:sim2sim}
\subsubsection{Cross-Engine Simulation Environment}
\label{subsec:sim2sim_env}
To assess the transferability and robustness of the learned recovery
policy beyond its training distribution, we evaluate the frozen policy in
the Gazebo physics engine, driven by the same ROS\,2 architecture as the
physical vehicle and topped with an AprilTag detection node in place of a
ground-truth target feed, so that the perception--control loop matches the
on-robot deployment. The vehicle dynamics and thruster model follow the
same formulation as in Isaac~Sim, with the hydrodynamic parameters fixed at
the most adverse end of their training randomization range (weaker
self-stabilization and lower water damping).

To better approximate underwater visual conditions, the detection node is
subject to four perturbations: additive zero-mean Gaussian noise of
$3\,\mathrm{mm}$ ($1\sigma$) on the three-axis target-center position; a
$5\%$ per-frame solve-failure rate (emulating sporadic frame loss from
suspended particulates); a $35\,\mathrm{ms}$ pipeline delay (matching the
measured end-to-end latency on the NX board); and random occlusion events
triggered with probability $0.15$ per second, each lasting
$1$--$5\,\mathrm{s}$.

\subsubsection{Evaluation Trajectories}
\label{subsec:sim2sim_traj}

To probe generalization, the vehicle tracks a variable-depth 3-D orbit never
seen during training (Table~\ref{tab:trajectories}): a horizontal circle of
radius $r=0.75\,\mathrm{m}$, modulated by a
vertical sinusoid ($A_{z}=0.3\,\mathrm{m}$, $k_{z}=2$) whose out-of-plane
motion stresses full three-axis tracking.

We evaluate two target conditions: \emph{static}, with the AprilTag held
fixed; and \emph{moving}, where it follows a Lissajous trajectory as in
training (Table~\ref{tab:trajectories}), with $A=(2.0,\,2.0,\,0.6)\,\mathrm{m}$,
$\omega_{i}=2\pi\!\cdot\!0.01\,\mathrm{rad/s}$, and
$\varphi=(-1.6,\,1.8\pi,\,0.8\pi)\,\mathrm{rad}$. For moving-target trials the
occlusion duration is shortened to $1$--$3\,\mathrm{s}$; other perturbations
are unchanged.

\begin{table}[t]
\centering
\caption{Reference orbit and target trajectories.}
\label{tab:trajectories}
\renewcommand{\arraystretch}{1.3}
\footnotesize
\begin{tabular}{@{}ll@{}}
\toprule
\textbf{Trajectory} & \textbf{Parametrization} \\
\midrule
Ellipse &
$x = x_0,\ y = A_y\cos\varphi,\ z = z_0 + A_z\sin\varphi$ \\
Figure-8 &
$x = x_0 + A_z\sin 2\varphi\sin\theta,\ y = A_y\cos\varphi,$ \\
& $z = z_0 + A_z\sin 2\varphi\cos\theta$ \\
3-D orbit &
$x = r\cos\varphi,\ y = r\sin\varphi,\ z = z_0 + A_z\sin k_z\varphi$ \\
Lissajous (tgt) &
$p^{\,\mathrm{tgt}}_i = A_i\sin(\omega_i t+\varphi_i),\ i\in\{x,y,z\}$ \\
\bottomrule
\end{tabular}
\end{table}

\subsubsection{Baselines}
\label{subsec:sim2sim_baselines}

We compare four methods to assess their performance under the primary
metrics of target alignment and tracking accuracy:
\begin{itemize}
  \item \textbf{PBVS-PID:} a position-based visual servoing controller with a
        cascaded outer-loop reference generation and inner-loop PID tracking
        structure, in which vertical and horizontal motion are decoupled; the
        pitch and yaw channels keep the vehicle aligned with the tag center,
        while the horizontal and depth channels track the reference orbit trajectory.
  \item \textbf{PBVS-PID~+~Rec:} the same PID augmented with a hand-designed
        recovery module that, upon detecting a failed measurement, resets
        the reference orbit trajectory to zero while holding the attitude
        and depth of the last visible frame.
  \item \textbf{AquaOrbit (w/o Rec):} our policy architecture trained \emph{without}
        dropout-segment observations or the explicit recovery reward.
  \item \textbf{AquaOrbit:} the full proposed policy (this work).
\end{itemize}
For both recovery-free variants (PBVS-PID and AquaOrbit (w/o Rec)), the observation
during a lost frame is carried over from the previous frame.

For each method and target condition we run $20$ continuous orbiting
trials; a cycle is terminated early and counted as a failure if the range
to the AprilTag exceeds $3\,\mathrm{m}$, or if detection is continuously
lost for more than $5\,\mathrm{s}$ outside any scripted occlusion, after
which it is restarted. We report the completion rate, path deviation, and
radius error $e_R$, together with the \textbf{LOS error} (the absolute
line-of-sight bearing error over frames with a valid detection) and the
\textbf{attitude stability}, the time-averaged magnitude of the
composite three-axis angular acceleration.

\begin{figure*}[t]
      \centering
      \begin{minipage}{0.164\textwidth}
        \centering
        \includegraphics[width=\linewidth]{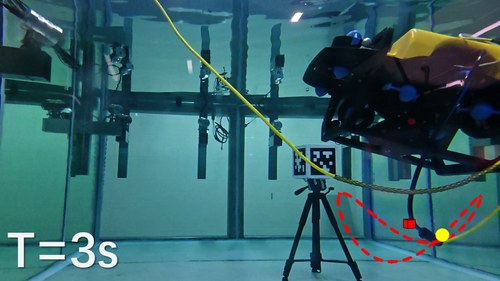}
      \end{minipage}\hfill
      \begin{minipage}{0.164\textwidth}
        \centering
        \includegraphics[width=\linewidth]{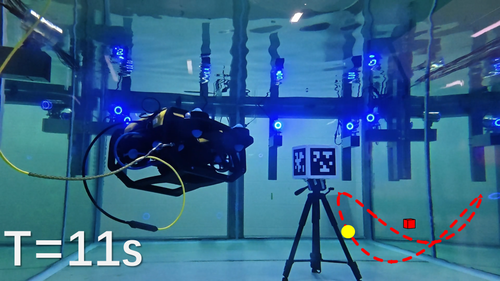}
      \end{minipage}\hfill
      \begin{minipage}{0.164\textwidth}
        \centering
        \includegraphics[width=\linewidth]{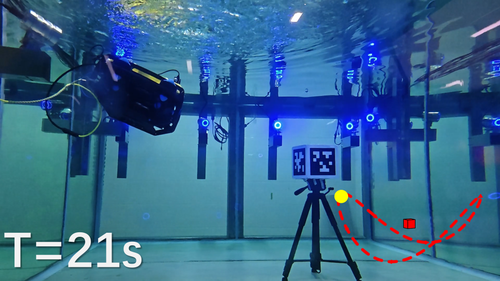}
      \end{minipage}\hfill
      \begin{minipage}{0.164\textwidth}
        \centering
        \includegraphics[width=\linewidth]{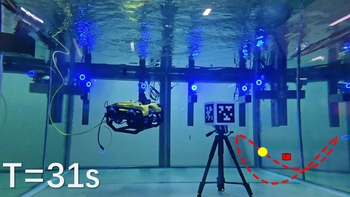}
      \end{minipage}\hfill
      \begin{minipage}{0.164\textwidth}
        \centering
        \includegraphics[width=\linewidth]{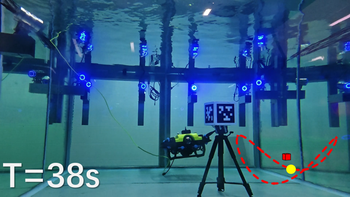}
      \end{minipage}\hfill
      \begin{minipage}{0.164\textwidth}
        \centering
        \includegraphics[width=\linewidth]{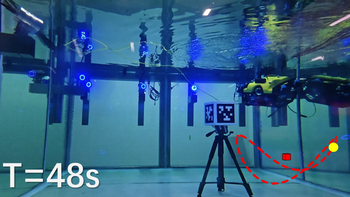}
      \end{minipage}

      \vspace{1mm}
      \begin{minipage}{0.164\textwidth}
        \centering
        \includegraphics[width=\linewidth]{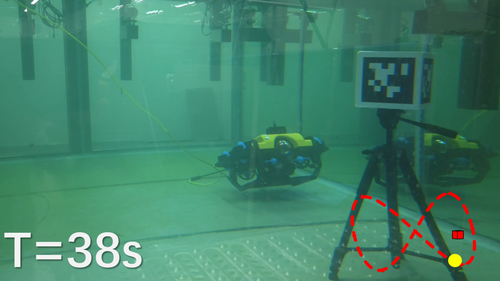}
      \end{minipage}\hfill
      \begin{minipage}{0.164\textwidth}
        \centering
        \includegraphics[width=\linewidth]{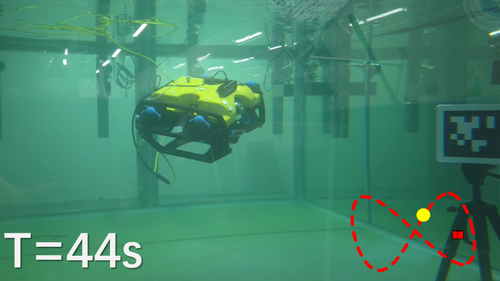}
      \end{minipage}\hfill
      \begin{minipage}{0.164\textwidth}
        \centering
        \includegraphics[width=\linewidth]{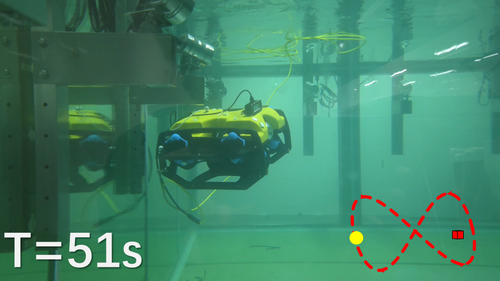}
      \end{minipage}\hfill
      \begin{minipage}{0.164\textwidth}
        \centering
        \includegraphics[width=\linewidth]{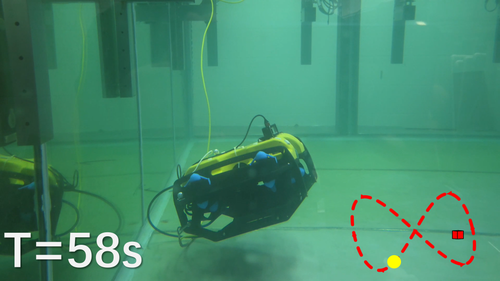}
      \end{minipage}\hfill
      \begin{minipage}{0.164\textwidth}
        \centering
        \includegraphics[width=\linewidth]{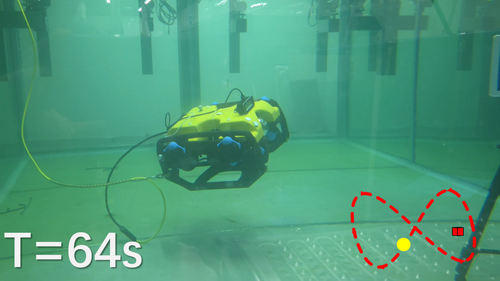}
      \end{minipage}\hfill
      \begin{minipage}{0.164\textwidth}
        \centering
        \includegraphics[width=\linewidth]{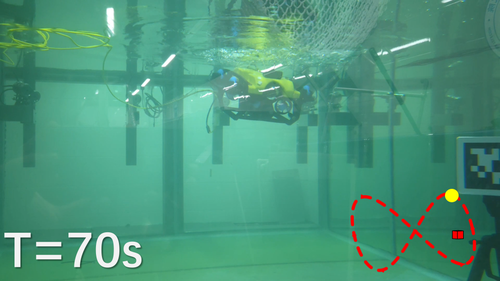}
      \end{minipage}

      \vspace{1mm}
      \begin{minipage}{0.164\textwidth}
        \centering
        \includegraphics[width=\linewidth]{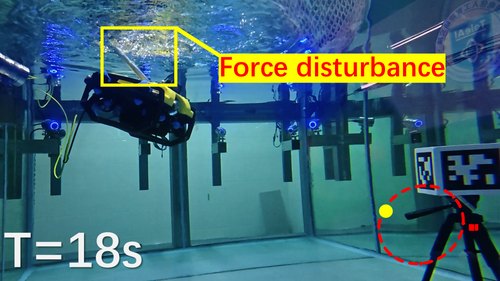}
      \end{minipage}\hfill
      \begin{minipage}{0.164\textwidth}
        \centering
        \includegraphics[width=\linewidth]{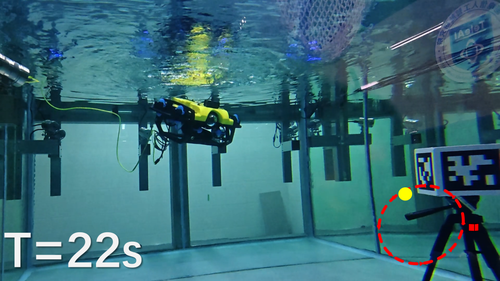}
      \end{minipage}\hfill
      \begin{minipage}{0.164\textwidth}
        \centering
        \includegraphics[width=\linewidth]{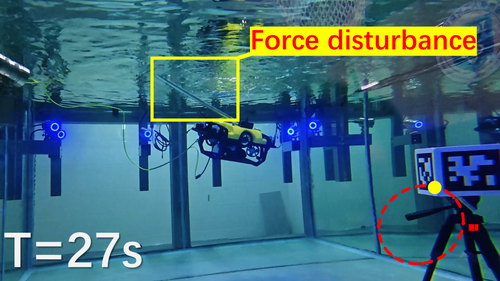}
      \end{minipage}\hfill
      \begin{minipage}{0.164\textwidth}
        \centering
        \includegraphics[width=\linewidth]{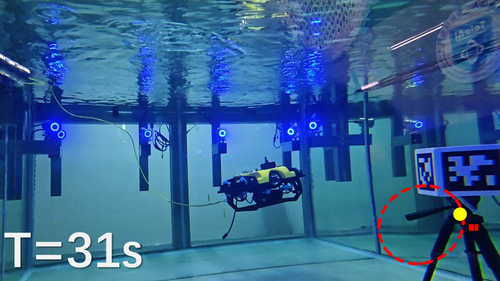}
      \end{minipage}\hfill
      \begin{minipage}{0.164\textwidth}
        \centering
        \includegraphics[width=\linewidth]{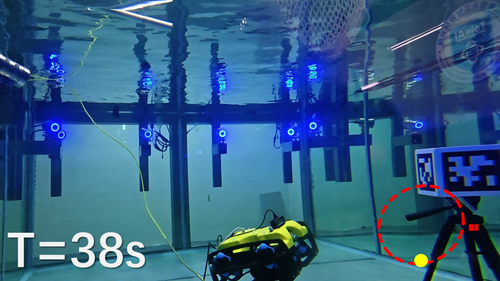}
      \end{minipage}\hfill
      \begin{minipage}{0.164\textwidth}
        \centering
        \includegraphics[width=\linewidth]{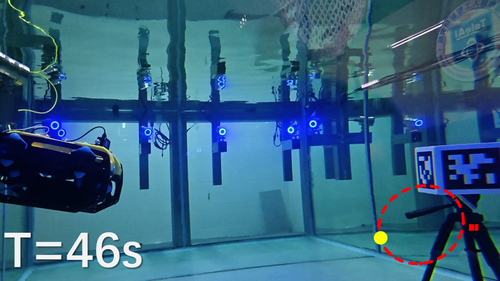}
      \end{minipage}

      \vspace{1mm}

      \begin{minipage}{0.332\textwidth}
        \centering
        \includegraphics[width=\linewidth]{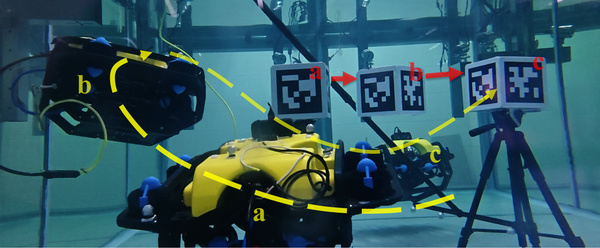}
      \end{minipage}\hfill%
      \begin{minipage}{0.332\textwidth}
        \centering
        \includegraphics[width=\linewidth]{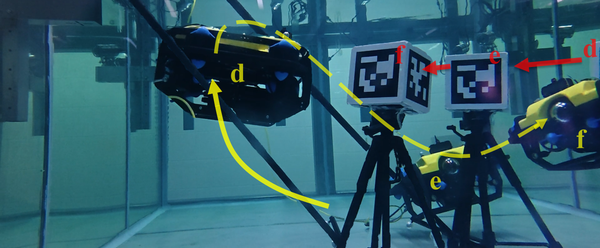}
      \end{minipage}\hfill%
      \begin{minipage}{0.332\textwidth}
        \centering
        \includegraphics[width=\linewidth]{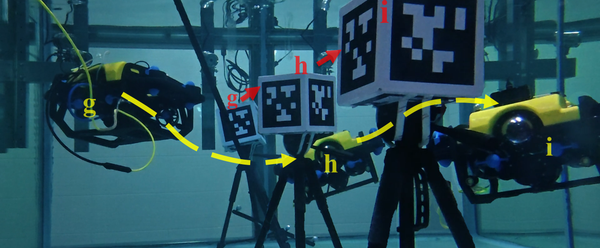}
      \end{minipage}
      \vspace{0mm}
      \caption{Rows, top to bottom: 3D orbit; $30^\circ$-pitch Figure-8 (water
      quality was relatively poor here); vertical-plane Ellipse under applied
      force disturbances; and manually controlled irregular AprilTag motion
      (multi-directional linear/curved motions and box-shaped attitude
      oscillations).}
      \label{fig:relpos}
  \vspace{-3mm}
  \end{figure*}
  
\subsubsection{Results}
\label{subsec:sim2sim_results}

Table~\ref{tab:sim2sim} summarizes the quantitative comparison.
AquaOrbit completes all $20$ continuous orbiting cycles under both the
static- and moving-target conditions---the highest success rate among all
methods---while attaining the lowest line-of-sight error; its
position-tracking accuracy is on par with the strongest baseline. In the dynamic scenario, PBVS-PID~+~Rec attains competitive
path deviation and attitude stability, but its LOS error is markedly
higher than ours, and its success rate on the moving-target task drops to
$18/20$; the failures occur when the relative velocity with respect to
the tag is large and the orbiting path is near the extremal points of its
depth profile.
The contribution of the recovery mechanism is more pronounced in the
dynamic-target scenario: removing the recovery module reduces the
completion rate of the reinforcement-learning policy from $20/20$ to
$9/20$, and that of the PID method from $18/20$ to $5/20$. As the
occlusion duration grows (beyond roughly $2\,\mathrm{s}$), both
recovery-free methods exhibit sharply higher loss rates accompanied by a
degradation in LOS tracking accuracy. This stems from their reliance on
coasting on stale measurements during visual dropout, followed by large
abrupt attitude and position corrections upon reacquisition---an effect
that is more severe for the PID-based method.

\begin{table}[t]
\centering
\caption{Cross-engine comparison and ablation on the 3-D orbit.}
\label{tab:sim2sim}
\setlength{\tabcolsep}{3.5pt}
\renewcommand{\arraystretch}{1.15}
\resizebox{\columnwidth}{!}{%
\begin{tabular}{llccccccc}
  \toprule
  \multirow{2}{*}{Target} & \multirow{2}{*}{Method}
  & \multirow{2}{*}{\shortstack{Succ.\\rate}}
  & \multicolumn{2}{c}{LOS err.\ [$^{\circ}$]}
  & \multicolumn{2}{c}{Path dev.\ [m]}
  & \multirow{2}{*}{\shortstack{$e_R$\\{[}m{]}}}
  & \multirow{2}{*}{\shortstack{$|\alpha|$\\{[}rad/s$^2${]}}} \\
  \cmidrule(lr){4-5}\cmidrule(lr){6-7}
  & & & Mean & P90 & Mean & RMS & & \\
  \midrule
\multirow{4}{*}{Static}
& PBVS-PID           & 12/20 & 9.64 & 13.08 & 0.145 & 0.186 & 0.069 & 1.205 \\
& PBVS-PID~+~Rec     & \textbf{20/20} & 9.49 & 13.10 & \textbf{0.100} &
\textbf{0.107} & 0.051 & \textbf{0.772} \\
& AquaOrbit (w/o Rec) & 18/20 & 12.65 & 18.57 & 0.196 & 0.214 & 0.089 & 0.793 \\
& \textbf{AquaOrbit} & \textbf{20/20} & \textbf{5.24} & \textbf{8.66} & 0.125
& 0.144 & \textbf{0.045} & 0.877 \\
\midrule
\multirow{4}{*}{Moving}
& PBVS-PID           & 5/20  & 10.50 & 15.29 & 0.168 & 0.179 & 0.135 & 0.969 \\
& PBVS-PID~+~Rec     & 18/20 & 9.64 & 13.89 & \textbf{0.139} & \textbf{0.153} &
0.113 & \textbf{0.888} \\
& AquaOrbit (w/o Rec) & 9/20  & 9.75 & 17.97 & 0.224 & 0.253 & 0.145 & 1.325 \\
& \textbf{AquaOrbit} & \textbf{20/20} & \textbf{5.19} & \textbf{9.68} &
0.141 & 0.157 & \textbf{0.088} & 1.141 \\
\midrule
\multirow{4}{*}{\shortstack[l]{Moving\\(Ablation)}}
& w/o $\Delta$Obs    & 19/20 & 5.92 & 9.89 & 0.171 & 0.221 & 0.106 & 1.080 \\
& w/o Jitter         & 16/20 & 7.52 & 11.62 & 0.166 & 0.180 & 0.136 & 2.094 \\
& w/o Latency        & 18/20 & 6.28 & 10.32 & 0.196 & 0.239 & 0.104 & 1.183 \\
& w/o ResetRand      & 16/20 & 6.52 & 10.77 & 0.231 & 0.256 & 0.144 & 1.238 \\
\bottomrule
\end{tabular}%
}
\end{table}

\subsubsection{Ablation Study}
\label{subsec:ablation}
With the experimental conditions unchanged, we ablate four factors in
isolation (Table~\ref{tab:sim2sim}, lower block): the historical
target-displacement observation (\textbf{w/o $\Delta$Obs}), the anti-jitter
terms in the reward and observation (\textbf{w/o Jitter}), the visual
measurement latency applied during training (\textbf{w/o Latency}), and the
reset randomization of the robot and reference orbit trajectory (\textbf{w/o
ResetRand}). As shown in the table, removing the anti-jitter terms affects
motion smoothness the most---$|\alpha|$ nearly doubles as short-period
pitch/yaw oscillation reappears; removing the reset randomization mainly
enlarges the path deviation and the LOS error, reflecting degraded
generalization to the tracking trajectory; the visual-latency and
$\Delta$Obs terms have milder effects, the latter concentrated on the LOS
error under the moving-target condition.

\begin{figure}[t]
\centering
\includegraphics[width=0.33\columnwidth]{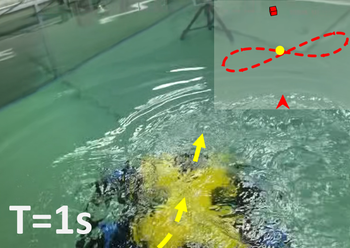}\hfill
\includegraphics[width=0.33\columnwidth]{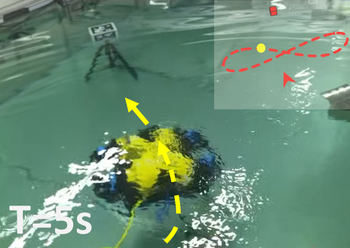}\hfill
\includegraphics[width=0.33\columnwidth]{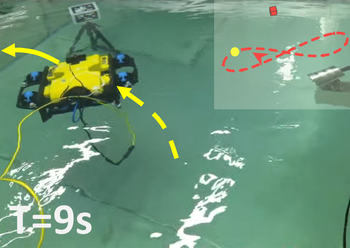}
\caption{The robot freely cuts into the reference orbit from an off-orbit release 
  point with no predefined guidance path, requiring only that the target tag 
  stay within the FOV.}
\label{fig:free-cutin}
\vspace{-3mm}
\end{figure}

\subsection{Sim-to-real Results}
\label{sec:results}
\subsubsection{Experimental Platform}
The experiments use a custom underwater robot with eight thrusters providing
full six-degree-of-freedom (6-DoF) actuation and a total mass of about 16~kg.
An NVIDIA Jetson~NX and a Pixhawk flight controller serve as the onboard
controllers: the Pixhawk receives the eight thruster commands from the
reinforcement-learning (RL) node and provides IMU and depth measurements, while
a monocular grayscale image serves as the visual input. All physical experiments were
conducted in a transparent experimental pond measuring $5~\mathrm{m} \times
3~\mathrm{m} \times 1.8~\mathrm{m}$ (length $\times$ width $\times$ water
depth). A tether transmits only start/stop commands; all perception, policy
inference, and control run entirely onboard, with no external computing devices
or sensors in the loop.
\begin{table}[t]
\centering
\caption{Real-world task evaluation across different scenarios}
\label{tab:real_world_eval}
\setlength{\tabcolsep}{3pt}
\renewcommand{\arraystretch}{1.15}
\resizebox{\columnwidth}{!}{%
\begin{tabular}{lccccccccc}
  \toprule
  \multirow{2}{*}{Trajectory} & \multirow{2}{*}{\shortstack{Succ.\\rate}}
  & \multicolumn{2}{c}{LOS err.\ [$^{\circ}$]}
  & \multicolumn{2}{c}{Path dev. (m)}
  & \multicolumn{3}{c}{Ang. accel. (rad/s$^2$)} \\
  \cmidrule(lr){3-4} \cmidrule(lr){5-6} \cmidrule(lr){7-9}
  & & Mean & P90 & Mean & RMS & $|\alpha_x|$ & $|\alpha_y|$ & $|\alpha_z|$ \\
  \midrule
Ellipse            & 7/8 & 11.96 & 17.89 & 0.150 & 0.169 & 0.272 & 0.227
& 0.130 \\
Figure-8           & 9/11 & 9.08  & 15.53 & 0.136 & 0.151 & 0.349 & 0.338
& 0.337 \\
3D Orbit (Occl.)   & 5/6 & 6.63  & 14.97 & 0.134 & 0.148 & 0.284 & 0.230
& 0.312 \\
3D Orbit (Mov.)    & 8/10 & 7.99  & 15.17 & 0.156 & 0.174 & 0.285 & 0.222
& 0.339 \\
\bottomrule
\end{tabular}%
}
\end{table}
\subsubsection{Real-World Pond Experiments}
We run multi-scenario experiments covering the three reference orbits (Ellipse,
Figure-8, 3D orbit; see Fig.~\ref{fig:relpos} and the video). On the 3D-orbit
task (Sec.~\ref{sec:static} and Sec.~\ref{sec:moving}), we focus on evaluating
the policy's recovery under sustained target occlusion and its ability to track
an irregularly moving AprilTag.

\subsubsection{Sustained-Occlusion Experiments}
\label{sec:static}
We emulate underwater moving obstacles by manually blocking the camera's line of
sight with a baffle (top of Fig.~\ref{fig:visionloss}). Without occlusion, the
3D-orbit period is about $60\,\mathrm{s}$ and the pitch
oscillates within $\pm 20^\circ$, as required to keep the target centered.

Long occlusions of $8\,\mathrm{s}$ and $5\,\mathrm{s}$ are introduced around
$75\,\mathrm{s}$ and $125\,\mathrm{s}$, and brief occlusions of under
$1\,\mathrm{s}$ around $23\,\mathrm{s}$, $115\,\mathrm{s}$, and
$139\,\mathrm{s}$. During occlusion the robot's yaw stays nearly unchanged and
its pitch and roll fluctuate within $2^\circ$, indicating a stable attitude.
Around $32\,\mathrm{s}$ and $87\,\mathrm{s}$ the target is lost to a large yaw
excursion rather than occlusion; the model re-acquires it within
$2.5\,\mathrm{s}$ by quickly adjusting its attitude.

Interestingly, the model learns a freer 3D motion in which the roll angle varies
periodically with the orbiting phase rather than being held at zero. This
emerges as a trade-off among the attitude-amplitude constraint, the thrust
(energy) penalty, and the smoothness penalty on angular velocity and
acceleration, yielding overall smoother motion.

\subsubsection{Moving-Target Experiments}
\label{sec:moving}
\begin{figure}[t]
\centering
\includegraphics[width=\columnwidth]{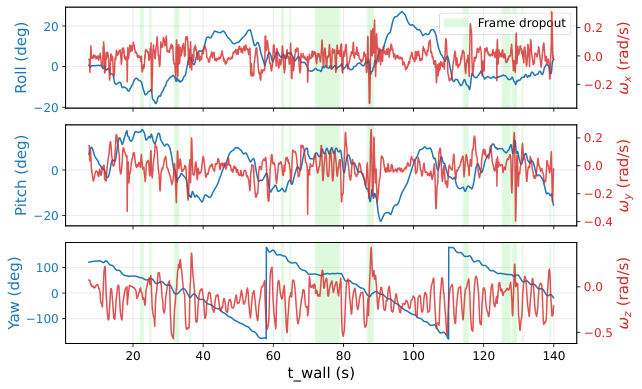}
\vspace{0.5mm}
\includegraphics[width=\columnwidth]{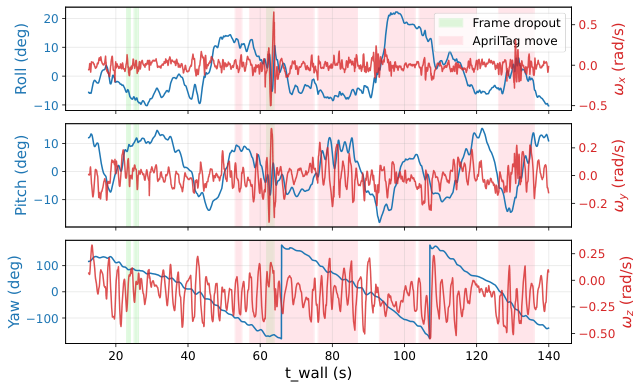}
\caption{Robot three-axis attitude angles and angular velocities over time.
  Top: the AprilTag is stationary throughout, with only manual occlusions.
  Bottom: the AprilTag moves during the pink-shaded intervals.
  $t_{\mathrm{wall}}=0$ denotes the instant of initial target acquisition.}
\label{fig:visionloss}
\end{figure}

We manually drive the AprilTag mount with a long rod, covering translations
of the AprilTag target along multiple directions, yaw rotations, and
attitude jitter, as shown in the bottom of Fig.~\ref{fig:visionloss}. Here, a single
occlusion of about $5\,\mathrm{s}$ is introduced at $23\,\mathrm{s}$, and
the AprilTag begins to move at $53\,\mathrm{s}$. The target loss lasting
about $2.5\,\mathrm{s}$ around $62\,\mathrm{s}$ is caused by the large
attitude jitter of the AprilTag during motion; in this case the robot
performs an aggressive maneuver and promptly recovers tracking. A clear
gradient in the robot's yaw is observed while the target is moving: for
example, the slope over $104$--$120\,\mathrm{s}$ is steeper than that over
the stationary-target interval of $120$--$125\,\mathrm{s}$, because the
target moves in the direction opposite to the robot's orbiting.
Fig.~\ref{fig:tracking} visualizes this task: when tracking the slowly moving
AprilTag, the policy attains position-tracking and line-of-sight alignment
accuracy comparable to the stationary-target case.

\subsubsection{Free Cut-in Without a Guidance Path}
\label{sec:fecutin}
\begin{figure}[t]
\centering

\begin{subfigure}[b]{0.60\columnwidth}
  \centering
  \includegraphics[width=\linewidth]{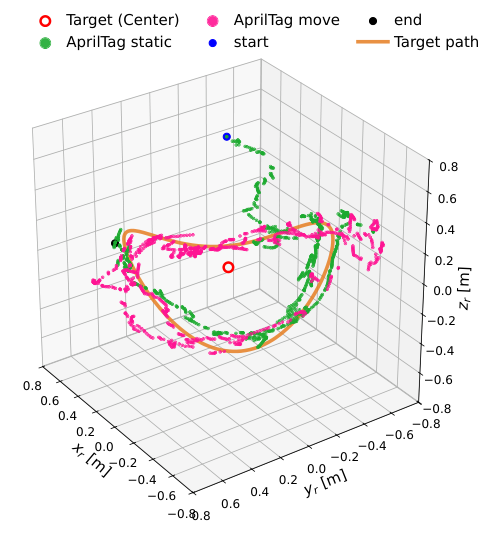}
  \label{fig:rel_orbit}
\end{subfigure}%
\hfill
\begin{subfigure}[b]{0.38\columnwidth}
\centering
\includegraphics[width=\linewidth]{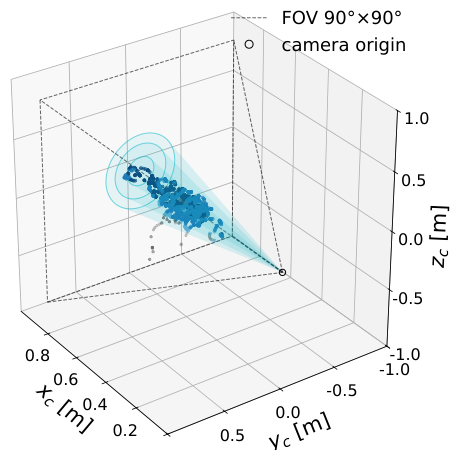}\\[3.0pt]
\includegraphics[width=\linewidth]{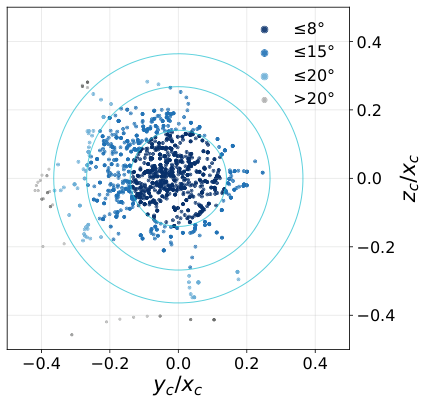}
\label{fig:relpos_right}
\end{subfigure}
\vspace{-5mm}
\caption{Left: 3D orbit of the ROV in the target frame; the ROV position is
reconstructed by fusing IMU measurements with visual target displacement.
Green/pink points denote samples with the AprilTag stationary/moving. Top-right:
3D vector from the camera to the AprilTag, with the dashed box indicating the
camera field of view (FOV). Bottom-right: perspective projection of the target
center on the image plane.}
\label{fig:tracking}
\end{figure}

None of the experiments above require the robot to be released on the reference
orbit with its line of sight pre-aligned to the tag. Fig.~\ref{fig:free-cutin}
shows the robot acquiring the tag from a large distance and merging into the
Figure-8 through a free arc, with no scripted guidance path; the left panel of
Fig.~\ref{fig:tracking} shows it launching from the water surface, descending
autonomously, and merging into the 3D orbit. Both cases show that AquaOrbit can
track from an off-orbit, sparse guidance point.

\subsubsection{Results Analysis}
Table~\ref{tab:real_world_eval} summarizes the pond experiments, with metrics
computed identically to the simulation. The real-world positional accuracy is
close to Gazebo and the line-of-sight error is slightly larger, while all
motion-jitter metrics fall below the simulated values, since the Gazebo
hydrodynamic coefficients and injected sensor noise were set to deliberately
aggressive levels. The Ellipse task shows a relatively higher LOS error, because
it involves larger in-plane displacement relative to the tag over most of the
cycle and incurs relatively more collisions with the wall edges. The
moving-target orbit exhibits the largest path deviation, $16\%$ above the
static-target case but still acceptable; its two terminations stem from target
loss caused by the tag approaching too quickly and from an accidental detachment
of the tag. The terminations on the Figure-8 and Ellipse both occur at the
lowest point of the trajectory, near the pond floor: the tag's lower side is
fixed to the mount so the ROV can only observe the tag's front face, and ground
contact induces abrupt pitch excursions that further raise the probability of
target loss. Overall, the policy exhibits strong cross-scenario generalization
in the real world and strong robustness to target jitter, external forces,
visual occlusion, and distant initial positions.

\section{Conclusion and Limitations}

\subsection{Conclusion}
We presented \method{}, an RL controller that maps visual
relative-position estimates, positional references, and onboard
state to thruster commands for underwater orbiting. Domain
randomization and a latched recovery mechanism enable operation
under intermittent visual feedback. Without retraining, in
Gazebo the policy completed 20/20 3-D orbiting trials with
random occlusion for both static and moving AprilTag targets;
removing recovery reduced moving-target completion to 9/20.
Compared with PBVS-PID with recovery, it reduced
moving-target line-of-sight error by 46\% with comparable
path accuracy. Zero-shot onboard deployment demonstrated
trajectory types unseen in training, free cut-in without a
predefined guidance path, attitude stability during occlusions
up to 8\,s, and reacquisition within 2.5\,s in reported
attitude-induced detection-loss events.

\subsection{Limitations and Future Work}
Validation relies on AprilTag-based position estimates and
pond experiments with manually driven targets. Recovery does
not predict target motion, and physical trajectories lack
independent ground truth. Future work will evaluate faster
targets, design a freer recovery strategy, improve measurement validation, and extend
the approach to natural targets and open water.

\bibliographystyle{IEEEtran}
\bibliography{ref}
\end{document}